\documentclass[runningheads]{llncs}
\usepackage[T1]{fontenc}

\usepackage{graphicx,verbatim, xcolor, multirow}
\usepackage{booktabs, amsmath, amssymb}

\begin{document}
\title{PolypSegTrack: Unified Foundation Model for Colonoscopy Video Analysis}
%
\begin{comment}  %% Removed for anonymized MICCAI 2025 submission
\author{First Author\inst{1}\orcidID{0000-1111-2222-3333} \and
Second Author\inst{2,3}\orcidID{1111-2222-3333-4444} \and
Third Author\inst{3}\orcidID{2222--3333-4444-5555}}
%
\authorrunning{F. Author et al.}
\institute{Princeton University, Princeton NJ 08544, USA \and
Springer Heidelberg, Tiergartenstr. 17, 69121 Heidelberg, Germany
\email{lncs@springer.com}\\
\url{http://www.springer.com/gp/computer-science/lncs} \and
ABC Institute, Rupert-Karls-University Heidelberg, Heidelberg, Germany\\
\email{\{abc,lncs\}@uni-heidelberg.de}}

\end{comment}

\author{Anwesa Choudhuri, Zhongpai Gao, Meng Zheng, Benjamin Planche, Terrence Chen, Ziyan Wu}  %% Added for anonymized MICCAI 2025 submission
\authorrunning{Choudhuri et al.}
\institute{United Imaging Intelligence, Boston, MA, USA \\
    \email{first.last@uii-ai.com}}

\maketitle              % typeset the header of the contribution
\begin{abstract}
Early detection, accurate segmentation, classification and tracking of polyps during colonoscopy are critical for preventing colorectal cancer. Many existing deep-learning-based methods for analyzing colonoscopic videos either require task-specific fine-tuning, lack tracking capabilities, or rely on domain-specific pre-training. In this paper, we introduce \textit{PolypSegTrack}, a novel foundation model that jointly addresses polyp detection, segmentation, classification and unsupervised tracking in colonoscopic videos. Our approach leverages a novel conditional mask loss, enabling flexible training across datasets with either pixel-level segmentation masks or bounding box annotations, allowing us to bypass task-specific fine-tuning. Our unsupervised tracking module reliably associates polyp instances across frames using object queries, without relying on any heuristics. We leverage a robust vision foundation model backbone that is pre-trained unsupervisedly on natural images, thereby removing the need for domain-specific pre-training. Extensive experiments on multiple polyp benchmarks demonstrate that our method significantly outperforms existing state-of-the-art approaches in detection, segmentation, classification, and tracking.


\end{abstract}

\section{Introduction}
% \acnote{Contributions:
% 1. joint training (detection, segmentation),
% 2. unsupervised tracking,
% 3. No need of domain-specific pre-training}

Early diagnosis of polyps in the gastrointestinal (GI) tract through colonoscopy is vital for preventing colorectal cancer. Automating the detection, segmentation, classification, and tracking of polyps in colonoscopic videos can greatly enhance the speed, accuracy, and consistency of polyp diagnosis. In recent years, deep learning methods~\cite{deformabledetr,querynet,unet,mask2former} has achieved remarkable progress in medical image analysis tasks such as segmentation and object detection. In particular, foundation models~\cite{sam2,clip}—pre-trained on large-scale, diverse datasets and then fine-tuned for specific tasks—have emerged as a promising direction for robust visual representation learning. Recent studies~\cite{dinov2} have shown that self-supervised learning in these models can yield general-purpose features that transfer well to downstream tasks.
Several works~\cite{endodino,endofm} have focused on developing foundation models for colonoscopic video analysis, specifically for polyp detection and segmentation. 

The aforementioned methods suffer from some drawbacks. The foundation models need to be fine-tuned separately on the different downstream tasks like detection and segmentation. However, treating these tasks independently neglects the inherent synergies between these tasks and limits the amount of data available for fine-tuning. Some recent works in computer vision~\cite{maskdino,openseed} have incorporated these synergies, but such synergies for colonoscopic video analysis is still lacking. Furthermore, the lack of large-scale video datasets with temporally dense segmentations has limited the development of effective tracking models for polyps, even though polyp tracking can be valuable for clinicians to generate exam/operation reports with accurate numbers and consistent labels of polyps. Additionally, the pre-training phase of the colonoscopic foundation models~\cite{endofm,endodino} relies on domain-specific data, i.e., colonoscopic videos, which can be expensive to collect. 

In this paper, we propose \textbf{PolypSegTrack}, a novel foundation model for polyp detection, segmentation, and unsupervised tracking in colonoscopic videos. Our novel conditional mask loss allows us to exploit the interdependencies between detection and segmentation tasks, and train our model in a flexible manner adapting to different annotation types. We develop an unsupervised and non-heuristic tracking approach that uses object queries to assign track identities to polyps. Our model is initially pre-trained on natural images in an unsupervised manner, which reduces the reliance on large-scale, expensive domain-specific colonoscopic data.

We evaluate our model on a wide range of tasks: joint detection and segmentation on the ETIS, CVC-ColonDB, CVC-300, Kvasir-SEG and the CVC-Clinic-DB datasets; detection and classification on the KUMC dataset; and joint detection and tracking on a subset of the REAL-Colon dataset. In all the aforementioned tasks, our model achieves the state-of-the-art results.

\begin{figure}[t]
    \centering
    \includegraphics[width=0.95\textwidth]{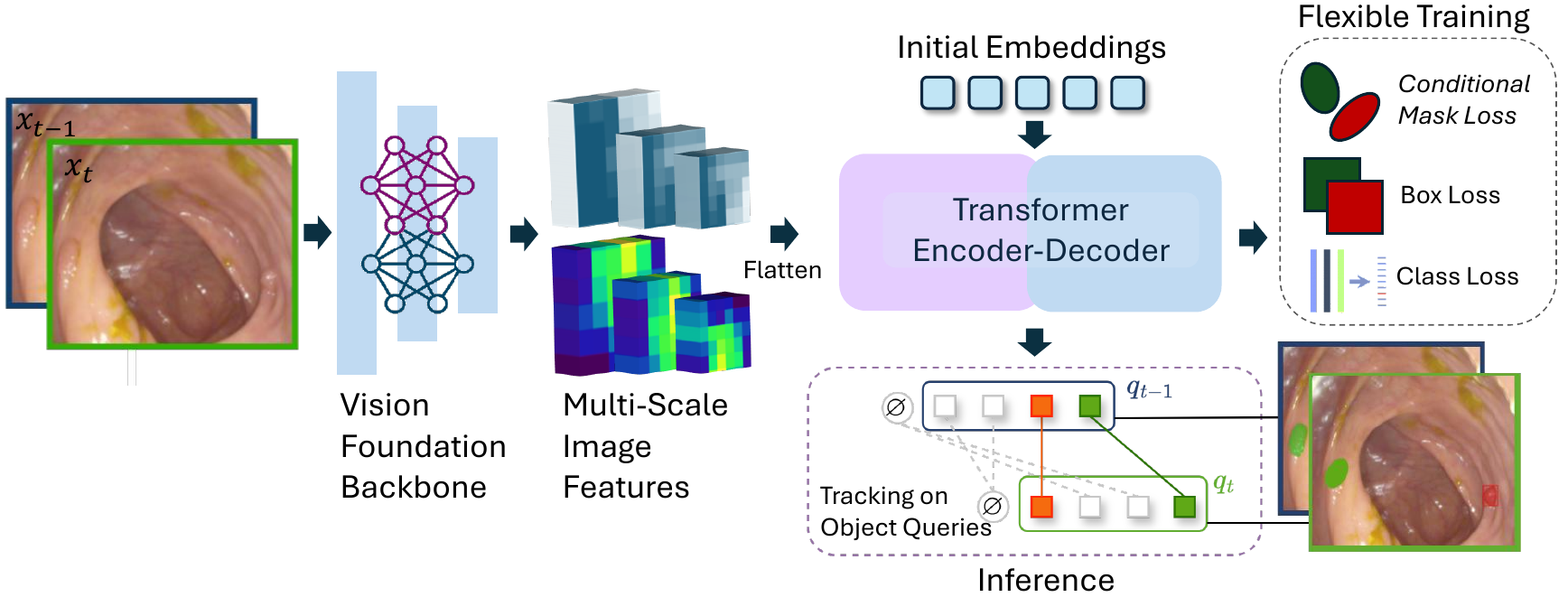}
    \caption{Overview of the proposed approach. Our proposed conditional mask loss (Sec.~\ref{sec:app:maskloss}) allows flexible training. Our unsupervised and non-heuristic tracking on object queries (Sec.~\ref{sec:app:tracking}) allows effective association of polyps across video frames.}
    \label{fig:mainfig}
\end{figure}

\section{Method}

% \subsection{Overview \zgnote{this title is optional}}
% \label{sec:app:overview}

Fig.~\ref{fig:mainfig} shows the overview of our approach. Given a video consisting of $T$ frames, the goal is to generate bounding boxes, segmentation masks, class probabilities, and track identities for every polyp in each video frame. Our model generates predictions in 2 stages: In stage 1, it produces bounding boxes, segmentation masks, and class probabilities of objects in each frame; and in stage 2, the objects are matched between every two consecutive frames to perform tracking. 

%\zgnote{This paragraph is optional depending on the page limit.} 
Sec.~\ref{sec:app:arch} describes stage 1 of our approach, i.e., joint detection, segmentation and classification in each frame. %Our choice of backbone allows us to leavarage large-scale unsupervised pre-training on natural images for domain-specific endoscopic images.
In Sec.~\ref{sec:app:maskloss}, we describe our novel conditional mask loss used to fine-tune our model jointly on training data containing segmentation masks, as well as on training data where only bounding box annotations are available. In Sec.~\ref{sec:app:tracking}, we describe stage 2 of our two-stage process, i.e., our non-heuristic and unsupervised tracking method during inference. 

\subsection{Joint Detection, Segmentation, and Classification}
\label{sec:app:arch}
To perform joint detection, segmentation and classification (stage 1 of our 2 stage process), our model consists of the following main components: a) a vision foundation model backbone to extract meaningful image features from every video frame, 
b) a transformer encoder and decoder to produce abstract object proposals or object queries, 
and c) prediction heads to produce the bounding boxes, segmentation masks and class probabilities using the object queries for each object.  %  The object queries play an important role for the success of an end-to-end pipeline like ours for set prediction with unknown number of outputs. The number of queries are selected as the maximum number of output instances of the model. During inference, a subset of queries predict $\varnothing$ outputs to dynamically adjust the number of valid outputs.

\noindent \textbf{Vision foundation model backbone.} Given a video frame $x_t \in \mathbb{R}^{H \times W}$, the vision foundation model backbone extracts meaningful multi-scale image features from the frame  (as shown in Fig.~\ref{fig:mainfig}). Here $H$ and $W$ refer to the height and width of the video frame. We use a pre-trained DINOv2~\cite{dinov2} backbone for the feature extraction, followed by four multi-layer perceptrons (MLPs) to generate features at multiple scales ($1/32$\textsuperscript{th}, $1/16$\textsuperscript{th}, $1/8$\textsuperscript{th} and $1/4$\textsuperscript{th}).  DINOv2~\cite{dinov2} is pre-trained on natural images using self-supervised learning and has shown to generate general-purpose features that transfer well to downstream tasks. Note that the lack of supervision during pre-training is important to capture visual features which translates well to other domains, like colonoscopic videos.

\noindent \textbf{Transformer encoder-decoder.} The transformer encoder-decoder accepts a set of trainable initial embeddings $e\in \mathbb{R}^{N \times C}$ and the flattened image features as inputs to produce $N$ object queries $q_t \in \mathbb{R}^{N\times C}$ for video frame $x_t$ (as shown in Fig.~\ref{fig:mainfig}). The object queries are abstract representations for all objects in the current frame. $N$ refers to the maximum number of objects to be discovered in the current frame and $C$ refers to the number of channels. We use MaskDINO (DETR with Improved Denoising Anchor Boxes)~\cite{maskdino} as our choice of encoder-decoder. MaskDINO is trained on natural-image datasets where both segmentation masks and bounding boxes are available, so the encoder-decoder captures the synnergies between the detection and segmentation tasks. However, note that MaskDINO is not extensively trained on data where segmentation masks are lacking, which is a common case for our setting, i.e., for colonoscopic videos.

\noindent \textbf{Prediction Heads.} There are three prediction heads: box head, classification head, and mask head (omitted in  Fig.~\ref{fig:mainfig} for clarity). The box head and the classification head accept the object queries and produce the bounding boxes and the class probabilities of the corresponding objects. The mask head accepts the object queries and generates some intermediate queries which are then multiplied with the image features and thresholded to produce the segmentation masks for individual objects.
Note that the object queries are expressive and contain enough information about the respective objects that they represent such that when they are passed through the respective heads, they are able to produce the desired bounding boxes, class probabilities, and segmentation masks for the objects they represent.

%Our architecture consists of a large vision foundation model backbone to ensure good feature extraction for each of the video frames. We have a transformer encoder-decoder that accepts the image features as input, along with some initial pre-trained object embeddings. The transformer encoder-decoder generates meaningful object queries for every frame, following prior work (cite). 

% \noindent 

\subsection{Training with Conditional Mask Loss} %Flexible Training
\label{sec:app:maskloss}

%Our model doesn't need any video-based training.
During training, our model can be conditioned to learn from either segmentation mask and bounding box annotations, or only from bounding box annotations if the segmentation masks aren't available in the dataset. This flexibility allows us to train jointly from a wide range of datasets containing either types of annotations. Datasets with segmentation-based annotations offer fine-grained pixel-level localization for the model, but large-scale segmentation datasets are lacking for colonoscopic videos. Datasets with bounding box-based annotations, on the other hand, are available more commonly, even though, they offer only course-grained localization in the form of $4$ points. To leverage learning from both kinds of datasets, we design a conditional mask loss which is activated only when segmentation annotations are available. 

 Specifically, the conditional mask loss $\mathcal{L_{\text{cond-mask}}}$ is the combination of dice loss $L_\text{dice}$ and mask-based cross entropy loss $L_\text{mask}$ whenever segmentation annotations are present. This loss is $0$ when only bounding box annotations are present. Formally, let $s^i_\text{gt}$ refer to the segmentation annotations for a given ground truth object $i$. If the segmentation annotations aren't present, $s^i_\text{gt}=\emptyset$. Let $K$ represent the number of ground truth objects in the current image. Following DETR~\cite{detr}, we first match the ground truth objects with the predicted object queries (some object queries remain unmatched, since $N>K$).  Let $\sigma_i$ represent a match between an object query with the ground truth object $i$.
 Then the conditional mask loss $\mathcal{L}_{\text{cond-mask}}$ for the given image is defined as follows.
\begin{equation}
\mathcal{L}_{\text{cond-mask}} = \sum_{i=1}^{K} \mathbb{I}_{\{{s^i_\text{gt}\neq\emptyset}\}} \left[ L_{\text{mask}}(\sigma_i) + L_{\text{dice}}(\sigma_i) \right]
\end{equation}

Our overall training objective is to minimize the a total multi-task loss function $\mathcal{L}$.    
\begin{equation}
    \mathcal{L} = \mathcal{L}_{\text{cls}} + \mathcal{L}_{\text{bbox}} + \mathcal{L}_{\text{cond-mask}}
\end{equation}

Here, $\mathcal{L}_\text{cls}$ refers to the cross entropy loss for class prediction and $\mathcal{L}_\text{bbox}$ is a combination of $L1$ loss and the generalized IoU loss calculated for the matched predicted objects in the current image and the ground truth following DETR~\cite{detr}.

\subsection{Unsupervised Tracking in the Space of Object Queries}
\label{sec:app:tracking}
In stage 2 of our two-stage approach, we temporally associate object instances between frames for tracking during inference (shown in Fig.~\ref{fig:mainfig}). In prior works on tracking~\cite{trackiou}, this step often involves heuristics like computing mask overlap, which may not generalize well in case of large camera motion or occlusions, both of which are common for colonoscopic videos. To avoid heuristic post processing, we match the object queries in the query space, following MinVIS~\cite{minvis}. Specifically, given two consecutive video frames $x_{t-1}$ and $x_t$, we obtain the set of object queries $q_{t-1}$ and $q_t$, as described in Sec.~\ref{sec:app:arch}. We perform tracking by using the Hungarian matching algorithm on a cost matrix $M\in\mathbb{R}^{N \times N}$, where every element is $M^{i,j}=S_\text{cosine}(q^i_{t-1}, q^j_t)$. Here, $S_\text{cosine}(q^i_{t-1}, q^j_t)$ represents the cosine similarity between the $i^\text{th}$ element in $q_{t-1}$ and $j^\text{th}$ element in $q_t$. Since the appearance of objects change gradually in a video, the object queries representing the same object only change slightly in consecutive frames, leading to a high similarity between these queries.

This approach of per-frame matching in the query-space is less affected by occlusions, as compared to directly matching masks or bounding boxes in the pixel space, because the object queries are not directly tied to the spacial positions of objects in each frame. Further, we do not need heuristics to handle the birth and death of object
instances in this framework. Since the number of queries ($N$) is set to a high limit, it is larger than the actual number of instances ($K$). So, there are queries that produce empty objects (represented by $\varnothing$ in Fig.~\ref{fig:mainfig}). The death of an object instance happens when its query is matched to such an empty query for more than five frames. If an object query is matched to an empty query for less than five frames, the query is carried forward and concatenated to the next frame's object queries. The birth of an instance is correctly handled if the matched query embeddings have been null before the actual birth of the object instance. Since the matching process does not need training, tracking can be performed in an unsupervised manner. The unsupervised tracking is particularly useful because of the lack of availability of densely annotated open-source colonoscopic videos with polyps.

% \subsection{Architecture}

\section{Experiments}

%Please add the following packages if necessary:
%\usepackage{booktabs, multirow} % for borders and merged ranges
%\usepackage{soul}% for underlines
%\usepackage{xcolor,colortbl} % for cell colors
%\usepackage{changepage,threeparttable} % for wide tables
%If the table is too wide, replace \begin{table}[!htp]...\end{table} with
%\begin{adjustwidth}{-2.5 cm}{-2.5 cm}\centering\begin{threeparttable}[!htb]...\end{threeparttable}\end{adjustwidth}
\begin{table}[t]\centering
\caption{Joint detection and segmentation performance on seen datasets.} 
\label{tab:seen}
\setlength{\tabcolsep}{4pt}
\scriptsize

\begin{tabular}{l|l|r|cccc|cccc}\toprule
& & &\multicolumn{4}{c|}{Kvasir-SEG} &\multicolumn{4}{c}{CVC-ClinicDB} \\
Type &Method & Venue &Dice &IoU &Pre. &Rec. &Dice &IoU &Pre. &Rec. \\
\midrule
\multirow{5}{*}{Det} 
&PraNet~\cite{pranet} &MICCAI’20 &89.1 &82.9 &- &- &89.4 &83.5 &- &- \\
&UACANet~\cite{uacanet} &ACM MM’21 &91.4 &86.1 &- &- &93.6 &88.9 &- &- \\
&SSFormer-L~\cite{ssformerl} &MICCAI’22 &92.2 &87.1 &- &- &90.7 &85.6 &- &- \\
&Polyp-PVT~\cite{polyppvt} &CAAI’23 &92.2 &86.9 &- &- &93.4 &88.4 &- &- \\
&PVT-CAS~\cite{pvtcascade} &WACV’23 &92.2 &87.2 &- &- &93.6 &88.9 &- &- \\
\midrule
\multirow{4}{*}{Seg}  
%&DETR~\cite{detr} &ECCV’20 &- &- &91.1 &84.3 &- &- &95.7 &98.5 \\
&Def. DETR~\cite{deformabledetr} &ICLR’21 &- &- &90.2 &76.0 &- &- &95.5 &94.1 \\
&DAB-DETR~\cite{dabdetr} &CVPR’22 &- &- &90.7 &80.2 &- &- &94.0 &92.6 \\
&DINO~\cite{dino} &ICLR’23 &- &- &90.2 &76.0 &- &- &95.5 &92.7 \\
\midrule
\multirow{2}{*}{Joint} &QueryNet~\cite{querynet} &MICCAI'24 &93.3 &88.3 &91.7 &82.6 &94.2 &89.4 &97.0 &97.0 \\
&\bf Ours & &\textbf{94.7} &\textbf{91.0} &\textbf{98.0} &\textbf{97.0} &\textbf{95.6} &\textbf{91.8} &\textbf{98.4} &\textbf{98.9} \\
\bottomrule
\end{tabular}
\end{table}

\begin{table}[t]\centering
\caption{Joint detection and segmentation performance on unseen datasets.}
\label{tab:unseen}
\scriptsize
\setlength{\tabcolsep}{2.5pt}
\begin{tabular}{l|l|cccc|cccc|cccc}\toprule
Type & Method &\multicolumn{4}{c|}{ETIS} &\multicolumn{4}{c|}{CVC-ColonDB} &\multicolumn{4}{c}{CVC-300} \\
& &Dice &IoU &Pre. &Re. &Dice &IoU &Pre. &Re. &Dice &IoU &Pre. &Re. \\
\midrule
\multirow{5}{*}{Det} 
&PraNet~\cite{pranet} &66.5 &58.1 &- &- &74.7 &66.1 &- &- &87.5 &79.7 &- &- \\
&UACANet~\cite{uacanet} &77.0 &69.0 &- &- &75.9 &68.7 &- &- &91.3 &85.1 &- &- \\
&SSFormer-L~\cite{ssformerl} &80.1 &72.8 &- &- &81.3 &73.5 &- &- &90.3 &83.8 &- &- \\
&Polyp-PVT~\cite{polyppvt} &78.1 &69.7 &- &- &81.3 &72.9 &- &- &89.8 &82.8 &- &- \\
&PVT-CAS~\cite{pvtcascade} &78.6 &70.8 &- &- &81.6 &73.5 &- &- &89.2 &82.3 &- &- \\
\midrule
\multirow{4}{*}{Seg} 
%&DETR~\cite{detr} &- &- &74.7 &75.3 &- &- &76.4 &80.8 &- &- &90.2 &91.7 \\
&Def. DETR~\cite{deformabledetr} &- &- &72.6 &70.2 &- &- &79.9 &82.6 &- &- &90.5 &91.8 \\
&DAB-DETR~\cite{dabdetr} &- &- &73.6 &71.2 &- &- &77.5 &78.2 &- &- &88.5 &90.0 \\
&DINO~\cite{dino} &- &- &71.3 &68.3 &- &- &77.5 &78.2 &- &- &91.7 &91.7 \\
\midrule
\multirow{2}{*}{Joint} 
& QueryNet~\cite{querynet} &81.9 &74.0 &74.9 &77.4 &82.8 &75.9 &83.5 &85.3 &92.0 &86.0 &91.8 &93.3 \\
& \bf Ours &\textbf{91.4} &\textbf{85.3} &\textbf{94.2} &\textbf{93.8} &\textbf{83.3} &\textbf{76.3} &\textbf{88.8} &\textbf{91.7} &\textbf{93.2} &\textbf{87.8} &\textbf{98.6} &\textbf{97.4} \\
\bottomrule
\end{tabular}
\end{table}

\subsection{Evaluation Datasets and Metrics}
\label{sec:exp:data}
\textbf{Datasets.} We evaluate the detection and segmentation performance of our model on five popular polyp datasets as benchmarks: CVC-ClinicDB~\cite{clinicdb}, Kvasir-SEG~\cite{kvasir}, CVC-ColonDB~\cite{colondb}, ETIS~\cite{etis} and CVC-300~\cite{cvc300}. We follow the same setting as PraNet~\cite{pranet}, that is, only 900 images from the Kvasir-SEG dataset and 550 images from the CVC-ClinicDB dataset are used for training, and the remaining images are used to test the learning ability of our method. The other 3 datasets are completely unseen during training and are used to test the generalizability of our method.
%Note that in this training, we have also used the ... datasets for fine-tuning our foundation models along with the aforementioned images from Kvasir-SEG and CVC-ClinicDB 
We also evaluate the detection and classification performance of polyps on the KUMC~\cite{kumc} validation dataset. To evaluate the unsupervised tracking consistency of our method,  we use 1000 consecutive frames from three videos from the REAL-Colon dataset~\cite{realcolon} without any fine-tuning. Note that there are currently no openly available datasets, to the best of our knowledge, to evaluate joint detection, segmentation, classification and tracking together, hence we evaluate on the aforementioned tasks to cover all the tasks.

\noindent \textbf{Metrics.} For joint detection and segmentation, we use the precision and recall metrics to evaluate the detection performance and the dice and IoU scores to measure the segmentation accuracy to be consistent with prior works ~\cite{querynet,pranet}. For detection and classification on KUMC~\cite{kumc} dataset, we report the F1 score following prior work~\cite{endofm}. To evaluate tacking, we report the object tracking metrics of DetA (detection accuracy)~\cite{hota}, AssA (association accuracy)~\cite{hota}, HOTA~\cite{hota}, MOTA (multi-object tracking accuracy) and IDF1 following prior works on multi-object tracking~\cite{trackformer}.

\subsection{Quantitative Results}

\noindent \textbf{Performance on joint detection and segmentation.}
 Tab.~\ref{tab:seen} shows the performance of recent models on the held-out validation images of the Kvasir-SEG and the CVC-ClonicDB datasets. Tab.~\ref{tab:unseen} shows the performance of these models on the unseen CVC-ColonDB, ETIS and CVC-300 datasets. We observe that our model outperforms other methods for all the datasets, sometimes by a large margin (as seen for the ETIS dataset).

\noindent \textbf{Performance on polyp detection and classification.}
We evaluate different methods on the KUMC dataset in Tab.~\ref{tab:kumc}. Our method outperforms the next best method, EndoFM~\cite{endofm}, which is a also foundation model, significantly.

\begin{table}[t]
\begin{minipage}[t]{0.37\hsize}
\centering
    \caption{Results on the KUMC~\cite{kumc} dataset.}\label{tab:kumc}
    \scriptsize
    \setlength{\tabcolsep}{4pt}
    \begin{tabular}{l|c}\toprule
        Method & F1 Score \\\midrule
        YOLOv4~\cite{yolov4} &57.2 \\
        FasterRCNN~\cite{fasterrcnn} &57.7 \\
        RetinaNet~\cite{retinanet} &59.0 \\
        SSD~\cite{ssd} &66.5 \\
        TimeSformer~\cite{timesformer} & 75.8 \\
        CORP~\cite{corp} & 78.2 \\
        FAME~\cite{fame} & 76.9 \\
        ProViCo~\cite{provico} & 78.6 \\
        VCL~\cite{vcl} & 78.1 \\
        ST-Adapter~\cite{stadapter} & 74.9 \\
        Endo-FM~\cite{endofm} & 84.1 \\
        \bf Ours & \textbf{91.1} \\
        \bottomrule
    \end{tabular}
\end{minipage}%
\hfill
\begin{minipage}[t]{0.57\hsize}
\centering
    \caption{Tracking results on a subset of the REAL-colon~\cite{realcolon} dataset.} 
    \label{tab:tracking}
    \scriptsize
    \setlength{\tabcolsep}{4pt}
    \begin{tabular}{l|ccccc}\toprule
        Method & DetA  & AssA & HOTA & MOTA & IDF1 \\\midrule
       % CAROQ~\cite{caroq} & 32.7 & 25.4 & 38.4 & 31.6 & 29.6 \\
        IoU & 57.7& 28.2& 39.5& 33.6& 37.0 \\
        \bf Ours & 57.7 &\bf 49.9 &\bf 53.2 &\bf 34.3& \bf 52.7 \\
        \bottomrule
    \end{tabular}
    
    \vspace{0.3cm} % Space between tables
    \caption{Ablation on the ETIS dataset with Resnet-50, Swin-Large and DINOv2 backbones.}
    \label{tab:backbone}
    \scriptsize
    \setlength{\tabcolsep}{5pt}
    \begin{tabular}{l|cccc}\toprule
        & Dice & IoU & Pre. & Re. \\\midrule
        \textbf{Ours} (R50) &82.5 &76.4 &83.9 &87.5 \\
        \textbf{Ours} (SwinL) &89.9 &83.3 &88.8 &91.7 \\
        \textbf{Ours} (DINOv2) &\textbf{91.4} &\textbf{85.3} &\textbf{94.2} &\textbf{93.8} \\
        \bottomrule
    \end{tabular}
\end{minipage}
\end{table}

\noindent \textbf{Tracking performance.}
% No other methods, to the best of our knowledge. We compare several tracking methods combined with our setup.
To the best of our knowledge, we haven't seen any methods performing tracking on polyps. To analyze the tracking performance of our method, we use a subset of the REAL-Colon dataset.
Tab.~\ref{tab:tracking} shows the comparison of our method with heuristic-based IoU matching for tracking (row 1). Note that, the detection results are generated using the same model and hence the detection accuracy (DetA) is exactly the same for both these methods.

\noindent \textbf{Effect of different backbones.}
Tab.~\ref{tab:backbone} shows the performance of our model with different backbones, Resnet-50, Swin-L, and DINOv2. We see that DINOv2, being a general-purpose foundation model, outperforms the other models. 

\subsection{Qualitative Results}
\begin{figure}[t]
    \centering
    \includegraphics[width=0.95\textwidth, trim={0 7.8cm 0cm 0cm},clip]{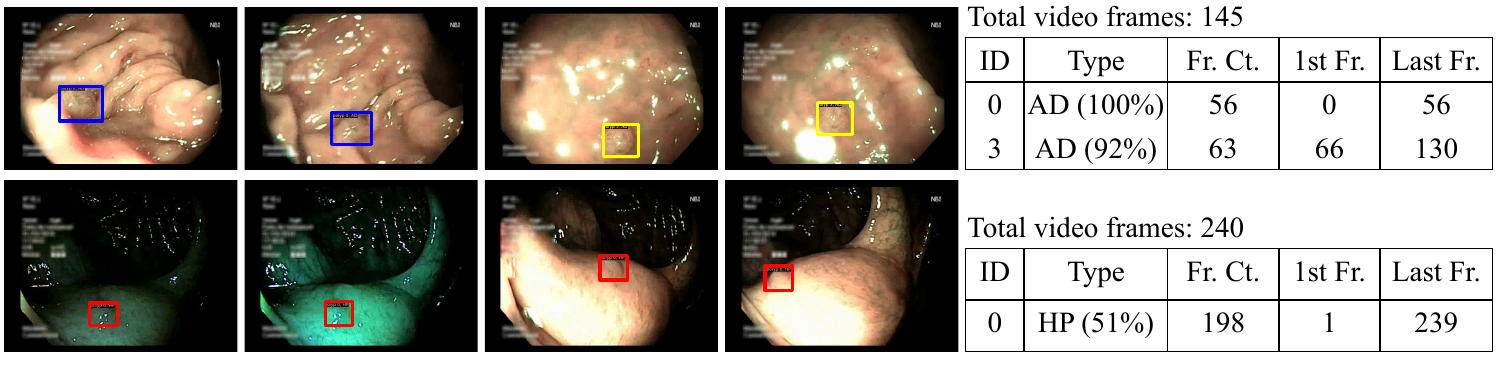}
    \caption{Detection, classification, and tracking on 2 videos (top and bottom row) of the KUMC dataset, along with comprehensive reports generated for each video. The reports summarize the polyps IDs, the polyp type (AD: cancerous or HP: benign) with prediction confidences, frame count (Fr. Ct.) of the polyps, their frame of first appearance (1st. Fr.) and their frame of last appearance (Last Fr.).}
    \label{fig:det_track_kumc}
\end{figure}

\begin{figure}[t]
    \centering
    \includegraphics[width=0.9\textwidth, trim={0 10.5cm 0cm 0cm},clip]{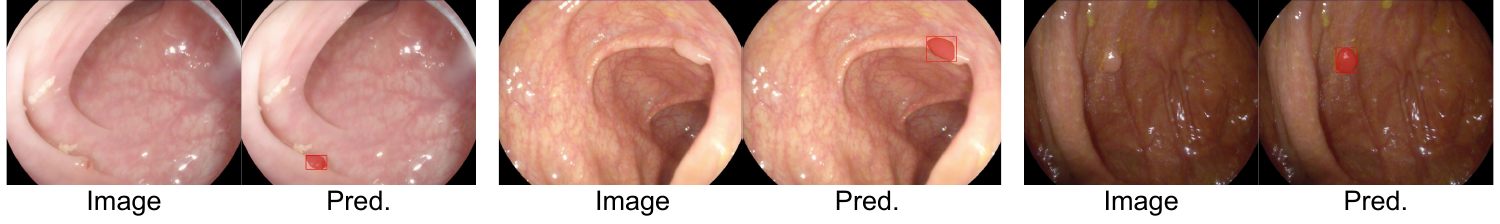}
    \caption{Detection and segmentation results on a few images from the ETIS dataset. Our model is able to discover small and hard to see polyps.}
    \label{fig:etis}
\end{figure}

Fig.~\ref{fig:det_track_kumc} shows two examples of joint detection, classification, and tracking on 2 videos of the KUMC dataset (top and bottom row), along with a comprehensive report generated for both videos. In the first example, we see 2 polyps (purple and yellow bounding boxes). Our model correctly identifies them as 2 different polyps. 
We see a similar example in the bottom row,  where our model is robust to different lighting conditions (green light and white light).
Fig.~\ref{fig:etis} shows three examples from the ETIS dataset where we perform joint detection and segmentation on small and hard-to-see polyps. Additional results are shown in the supplementary video.

\subsection{Training Data}
\label{sec:exp:impl}
For joint detection and segmentation (Tab.~\ref{tab:seen} and Tab.~\ref{tab:unseen}), our model is trained on 900 and 550 images of Kvasir-SEG and CVC-ClinicDB respectively, along with PolypDB~\cite{polypdb}, PolypGen~\cite{polypgen}, and KUMC~\cite{kumc} training datasets, to align our backbone and encoder-decoder to colonoscopic videos. This one-step fine-tuning removes the need to first pre-train our model on colonoscopic videos like in prior works with colonoscopic foundation models~\cite{endofm,endodino}. Note that KUMC~\cite{kumc} dataset only has bounding-box-based annotations (with 28k training images), whereas PolypDB, PolypGen have available segmentation masks (with 6k frames combined). The KUMC dataset has polyp classification information (i.e., whether a polyp is AD: cancerous or HP: non-cancerous). For this experiment, both classes are just treated as polyps. 
For detection and classification (Tab.~\ref{tab:kumc}), our model is directly fine-tuned on the KUMC~\cite{kumc} training dataset. 
For the unsupervised tracking experiment (Tab.~\ref{tab:tracking}), we directly use the trained model from joint detection and segmentation  (Tab.~\ref{tab:seen} and Tab.~\ref{tab:unseen}), and test it on the REAL-Colon data without fine-tuning. For Fig.~\ref{fig:det_track_kumc}, we directly use the tracking module on top of the model trained on the detection and classification task (Tab.~\ref{tab:kumc}).

\section{Conclusion}

In this paper, we introduced PolypSegTrack, a novel foundation model that jointly addresses
polyp detection, segmentation, classification and unsupervised tracking in colonoscopic videos. Our novel conditional mask loss enables flexible training and our unsupervised and non-heuristic tracking approach reliably tracks polyp instances across video frames. Extensive experiments on multiple polyp benchmarks demonstrate that our method significantly outperforms existing state-of-the-art approaches in polyp detection,
segmentation, classification and tracking.


\bibliographystyle{splncs04}
\bibliography{ref}

\end{document}